\documentclass[conference]{IEEEtran}
\IEEEoverridecommandlockouts
\usepackage{cite}
\usepackage{amsmath,amssymb,amsfonts}
\usepackage{algorithmic}
\usepackage{graphicx}
\usepackage{textcomp}
\usepackage{xcolor}
\usepackage{multirow}
\usepackage{hyperref}
\def\BibTeX{{\rm B\kern-.05em{\sc i\kern-.025em b}\kern-.08em
    T\kern-.1667em\lower.7ex\hbox{E}\kern-.125emX}}

\begin{document}

\title{CLIP-HandID: Vision-Language Model for Hand-Based Person Identification \\
}

\author{\IEEEauthorblockN{Nathanael L. Baisa, Babu Pallam, Amudhavel Jayavel}
\IEEEauthorblockA{\textit{School of Computer Science and Informatics} \\
\textit{De Montfort University}\\
Leicester LE1 9BH, UK. \\
Email: nathanael.baisa@dmu.ac.uk.}
}

\IEEEoverridecommandlockouts


\maketitle

\IEEEpubidadjcol

\begin{abstract}

This paper introduces a novel approach to person identification using hand images, designed specifically for criminal investigations. The method is particularly valuable in serious crimes such as sexual abuse, where hand images are often the only identifiable evidence available. Our proposed method, CLIP-HandID, leverages a pre-trained foundational vision-language model - CLIP - to efficiently learn discriminative deep feature representations from hand images (input to CLIP's image encoder) using textual prompts as semantic guidance. Since hand images are labeled with indexes rather than text descriptions, we employ a textual inversion network to learn pseudo-tokens that encode specific visual contexts or appearance attributes. These learned pseudo-tokens are then incorporated into textual prompts, which are fed into CLIP's text encoder to leverage its multi-modal reasoning and enhance generalization for identification. Through extensive evaluations on two large, publicly available hand datasets with multi-ethnic representation, we demonstrate that our method significantly outperforms existing approaches. The source code is available at \textcolor{red}{\url{https://github.com/nathanlem1/CLIP-HandID}}.  
\end{abstract}

\begin{IEEEkeywords}
Person identification, Hand recognition, Multimodal deep learning, Vision-Language, CLIP model
\end{IEEEkeywords}

\section{Introduction} \label{sec:intro}

Biometric identification, which verifies individuals based on physical or behavioral traits, has gained significant interest for various applications. Among primary biometric traits, hand images provide distinctive features for person identification~\cite{DanEliRos16, BaiWilRahIPTA22, BaiWilRahICPR22, Baisajvci24, Baisaiwbf24}. Unlike other biometric modalities, hand images exhibit lower variability while maintaining robust and diverse characteristics that remain stable in adulthood~\cite{BaiWilRahIPTA22, BaiWilRahICPR22, YimChaShu20, AttAkhCha21}. This stability makes hand images a promising candidate for biometric analysis, particularly in forensic investigations where uncontrolled environments are common. In serious crimes like sexual abuse, hand images are often the only identifiable evidence available, underscoring their investigative value. 

Previous studies~\cite{Mah19,YimChaShu20} have explored hybrid approaches combining traditional machine learning and deep learning techniques for hand-based person identification. In~\cite{Mah19}, researchers trained a convolutional neural network (CNN) on RGB hand images and subsequently used the network as a feature extractor. The extracted CNN features were then classified using support vector machine (SVM) classifiers. Similarly,~\cite{YimChaShu20} adopted a comparable methodology but incorporated an additional data modality by fusing near-infrared (NIR) images with RGB data. However, these approaches lack a true end-to-end learning framework. End-to-end approaches have also been introduced in~\cite{BaiWilRahIPTA22, BaiWilRahICPR22, NicGonRat24}; however, they are based on uni-modal architectures and process only images.

Recently, vision-language cross-modal learning such as Contrastive Language-Image Pre-training (CLIP)~\cite{RadKimHal21}, which establishes meaningful connections between visual representations and their textual descriptions, has received great attention in deep learning. This model employs two key innovations: training on extensive datasets (image-text pairs) and reformulating the pre-training objective to align visual features with their linguistic counterparts using self-supervised contrastive learning, which minimizes cosine distance (maximizes cosine similarity) for matched image-text pairs. This paradigm enables the image encoder to capture richer semantic information from images with the help of textual prompts through text encoder and develop highly adaptable features that demonstrate strong transferability across diverse uni-modal and multi-modal downstream tasks~\cite{RenWeiRon22, MokHerBer21, HaoZhaWu23}, when compared to uni-modal CNN-based~\cite{HeZhaSun16} and Vision Transformer (ViT)-based~\cite{AleLucAle21} methods pre-trained on manually labelled images (ImageNet) from a set of predefined classes where visual information conveying semantic meaning beyond the predefined categories is totally disregarded. Consider a standard image classification scenario: the predefined class labels serve as concrete textual inputs that can be incorporated into descriptive prompts (e.g., "A photo of a [class]", where [class] is generally replaced by concrete text labels). The classification process occurs through cross-modal comparison - matching the visual features extracted from images through the image encoder against the textual features generated by processing these label-enhanced prompts through the text encoder - as a zero-shot solution during inference. This vision-language model has been adopted for person and/or vehicle re-identification in~\cite{LiSunLi23, YanWuWu24} where it has shown promising results. However, there is no work which has leveraged the vision-language foundation models such as CLIP for identification of a person based on hand images for criminal investigations. In fine-grained hand-based person identification, the labels are indexes (numeric values) instead of text descriptions. Therefore, we adopted textual inversion~\cite{RinYuvYuv23, BalAgnBer23} to describe the images with specific words to fully exploit the cross-modal description ability in CLIP for the development of hand-based person identification. This textual inversion network learns pseudo-tokens to generate customized textual descriptions for individual subjects. These pseudo-tokens enable the model to focus on semantically or meaningfully relevant visual regions that align precisely with the prompt context. This design allows the model to learn fine-grained attribute information.

In this work, we propose a hand-based person identification method, CLIP-HandID, by learning natural language guided deep feature representations from hand images captured by digital cameras. The proposed method follows an end-to-end training. Our contributions can be summarized as follows.
\begin{enumerate}
\item We design a novel method, CLIP-HandID, for hand-based person identification by leveraging vision-language multi-modal model, particularly CLIP. 
\item We learn pseudo-tokens that encode specific visual contexts using textual inversion network since hand images are labeled with indexes rather than text descriptions.
\item We make rigorous evaluations on two large multi-ethnic and publicly available hand datasets (11k hands~\cite{Mah19} and HD~\cite{KumXu16}).
\end{enumerate}

The rest of the paper is organized as follows. The proposed method including the overall architecture of CLIP-HandID is described in Section~\ref{sec:proposedMethod} followed the experimental results in Section~\ref{sec:experimentalResults}. The main conclusion along with suggestion for future work is summarized in Section~\ref{sec:Conclusion}.

\section{Proposed Method} \label{sec:proposedMethod}

\begin{figure*}[t]
\begin{center}
  \includegraphics[width=0.8\linewidth]{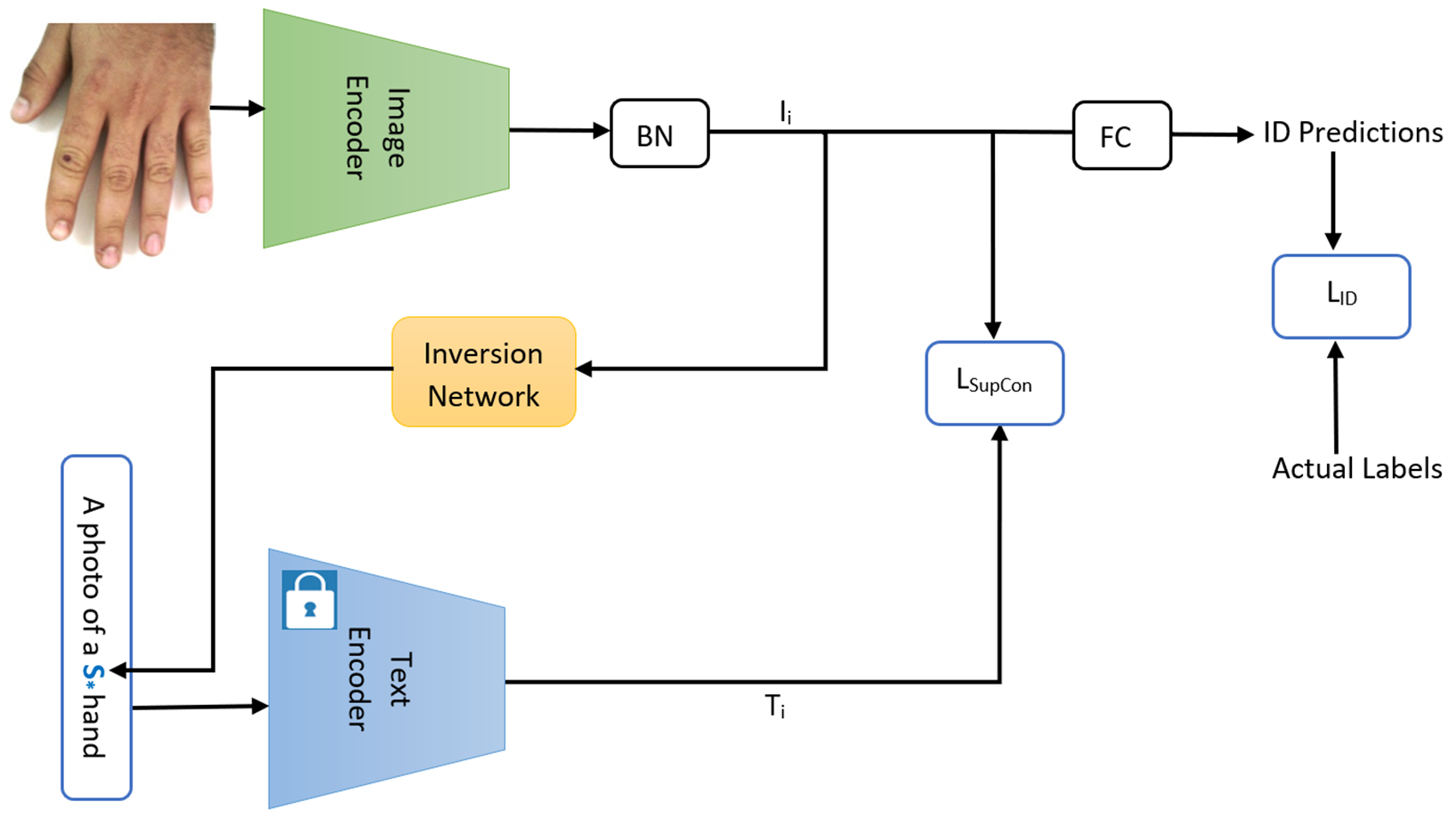} \\
\end{center}
   \caption{Structure of CLIP-HandID. Given an input image, visual embedding $I_i$ is obtained by passing it through the image encoder from the CLIP backbone network. The inversion network then inverts the visual embedding $I_i$ to a unique pseudo-word token $S_*$, which is then incorporated into a textual prompt as "A photo of a $S_*$ hand". The textual prompt is then given as input to the text encoder to obtain textual embedding $T_i$. The classifier predicts the identity (ID) of the input image during training. Note that only the image encoder is optimized while the text encoder is frozen.}
\label{fig:CLIP-HandID}
\end{figure*}

Our proposed method, CLIP-HandID, is summarized in Fig.~\ref{fig:CLIP-HandID}. We use CLIP~\cite{RadKimHal21} pre-trained on large collection of image-text pairs as a backbone network, which comprises two main components: an image encoder $\theta_I(.)$ and a text encoder $\theta_T(.)$. The image encoder offers architectural flexibility, supporting either a Vision Transformer (e.g. ViT-B/16) or a CNN backbone (e.g. ResNet-50), both capable of projecting images into a shared cross-modal embedding space. The text encoder employs a transformer architecture that processes input sentences following the pattern "A photo of a [class]", where [class] serves as a placeholder for specific textual labels. This text processing begins with lowercased byte pair encoding (BPE)~\cite{SenHadBir16} using a 49,152-token vocabulary, followed by conversion to 512-dimensional word embeddings. Each text sequence is standardized to 77 tokens (context length), including start ([SOS]) and end ([EOS]) tokens, and processed through 12 transformer layers with 8 attention heads each. The final text representation derives from the layer-normalized [EOS] token embedding, which is then projected into the shared embedding space.

For batch processing applications where $i \in \{1...B\}$ indexes images within a batch, the system computes similarity scores between two key representations: $I_i$, corresponding to the [CLS] token embedding from the image encoder, and $T_i$, representing the matching [EOS] token embedding from the text encoder. This dual-encoder framework enables effective cross-modal comparison within a common feature space.

While CLIP operates effectively in standard classification by reformulating labels as text prompts (yielding aligned visual $I_i$ and text $T_i$ embeddings), its direct application to hand-based person identification is problematic due to the non-semantic nature of indexes as hand image labels instead of text descriptions, which serve as mere instance identifiers. Therefore, we design textual inversion network to directly invert an image (visual embedding $I_i$) into a unique pseudo-token $S_*$, which can be incorporated into a textual prompt as "A photo of a $S_*$ hand" and then given as input to the text encoder of the CLIP, as shown in Fig.~\ref{fig:CLIP-HandID}. It is important to highlight that this pseudo-token does not correspond to any real word; instead, it serves as an abstract representation within the token embedding space. The learned pseudo-token effectively encode both global visual patterns and fine-grained visual characteristics of the original image. Furthermore, the learned pseudo-tokens by the textual inversion network are personalized textual descriptions for individual subjects which enable or guide the model to focus on semantically relevant visual regions (hand structure) that align precisely with the prompt context. We use a lightweight model for the inversion network consisting of three fully-connected (FC) layers with GELU activation function, layer normalization and dropout with a probability of 0.5 after each layer except the last layer. Only the layer normalization is used just after the last layer.

The visual embedding $I_i$ is given as input to a classification layer, which is implemented using a FC layer followed by a softmax function. We also include a batch normalization (BN) to reduce possible over-fitting, as shown in Fig.~\ref{fig:CLIP-HandID}. The classification layer predicts the identity (ID) of each input image. 

We optimize the CLIP-HandID during training by minimizing the sum of cross-entropy and supervised contrastive losses. Note that only the image encoder is optimized (or fine-tuned) while the text encoder is frozen. The cross-entropy loss is given as

\begin{equation}
    \mathcal{L}_{ID} = \sum_{k=1}^K -q_k \log(p_k)
\label{eqn:ID}
\end{equation}
\noindent where $K$ is the total number of classes (identities) and $p_{k}$ is the predicted probability of class (identity) $k$. We also use label smoothing~\cite{SzeVanIof16} which prevents over-fitting and over-confidence with smoothing value ($\epsilon$) of 0.1. The ground-truth distribution over labels $q_k$ by including label smoothing can be given as 

\begin{equation}
    q_k =
\begin{cases}
    1 - \frac{K-1}{K} \epsilon, & \text{if } k = y\\
    \frac{1}{K} \epsilon,              & \text{otherwise}
\end{cases}
\label{eq:labelSmoothing}
\end{equation}
\noindent where $y$ is ground-truth label.

The image encoder $\theta_I(.)$ takes an image $\mathbf{x}_i \in \mathbb{R}^{C \times H \times W}$ as input, where C, H and W are the number of channels, the height and the width, respectively.  Similarly, the text encoder $\theta_T(.)$ takes a tokenized textual prompt $\mathbf{t}_i  \in \mathbb{R}^{N \times D}$ as input, where N is the number of text sequences and D is the context length - the maximum number of tokens that the model can process in a single input sequence. Thus, the image feature $I_i = \theta_I(\mathbf{x}_i)$ and text feature $T_i = \theta_T(\mathbf{t}_i)$. Hence, to calculate supervised contrastive loss, we first need to compute the similarity between the embeddings of input image $\mathbf{x}_i$ and input tokenized textual prompt $\mathbf{t}_i$ as follows

\begin{equation}
    \text{sim}(I_i, T_i) = \text{sim}(\theta_I (\mathbf{x}_i), \theta_T (\mathbf{t}_i) )
\label{eqn:similarity}
\end{equation}
\noindent where $I_i = \theta_I (\mathbf{x}_i)$ and $T_i = \theta_T (\mathbf{t}_i)$ are the image feature and the text feature, respectively, projected into a cross-modal embedding space, and $\text{sim}(\mathbf{u},\mathbf{v}) = \frac{\mathbf{u} \cdot \mathbf{v}}{\|\mathbf{u}\|_2 \|\mathbf{v}\|_2}$ is the cosine similarity. Note that $\mathbf{u}\cdot\mathbf{v} = \mathbf{u}^T \mathbf{v}$. Then, the image-to-text contrastive loss $\mathcal{L}_{i2t}$ is calculated as:

\begin{equation}
    \mathcal{L}_{i2t}(i) = - \log \frac{\exp(\text{sim}(I_i, T_i)/\tau)}{\sum_{b=1}^B \exp(\text{sim}(I_i, T_b)/\tau)}
\label{eqn:i2t}
\end{equation}

and the text-to-image contrastive loss $\mathcal{L}_{t2i}$ is computed as:

\begin{equation}
    \mathcal{L}_{t2i}(i) = - \log \frac{\exp(\text{sim}(T_i, I_i)/\tau)}{\sum_{b=1}^B \exp(\text{sim}(T_i, I_b)/\tau)}
\label{eqn:t2i}
\end{equation} where the numerators in (\ref{eqn:i2t}) and (\ref{eqn:t2i}) represent the similarity scores between matched embedding pairs, while the denominators aggregate all similarity measures relative to the anchor $I_i$ or $T_i$, and $\tau$ represents the temperature, which we set $\tau=1$. Note that  $I_i$ and $T_i$ have the same dimension in the cross-modal embedding space.

Following CLIP's approach, we retain the $\mathcal{L}_{i2t}$ and $\mathcal{L}_{t2i}$ objectives but modify $\mathbf{t}_i$ to $\mathbf{t}_{y_i}$ in (\ref{eqn:similarity}) to account for shared text descriptions across each identity. Furthermore, since multiple images in a batch may correspond to the same person (hand), we adapt $\mathcal{L}_{i2t}$ and $\mathcal{L}_{t2i}$ to handle cases where $T_{y_i}$ can have multiple positive matches, with the goal of guiding the learned pseudo-token to capture distinctive visual details belonging to the same identity. The modified formulation becomes as follows by exploiting the symmetric supervised contrastive loss, obeying the principle of cycle-consistency:

\begin{equation}
    \mathcal{L}_{SupCon} = \mathcal{L}_{i2t}^{Sup} + \mathcal{L}_{t2i}^{Sup}
\label{eqn:supcon}
\end{equation}

\begin{equation}
    \mathcal{L}_{i2t}^{Sup} = - \frac{1}{|P(y_i)|} \sum_{p \in P(y_i)} \log \frac{\exp(\text{sim}(I_p, T_{y_i})/\tau)}{\sum_{b=1}^B \exp(\text{sim}(I_p, T_b)/\tau)}
\label{eqn:i2tsupcon}
\end{equation}

Here, $P(y_i) = \{p \in 1...B : y_p = y_i\}$ is the set of indices of all positives for $T_{y_i}$ in the batch while $|.|$ is its cardinality.

\begin{equation}
    \mathcal{L}_{t2i}^{Sup} = - \frac{1}{|P(y_i)|} \sum_{p \in P(y_i)} \log \frac{\exp(\text{sim}(T_{y_i}, I_p)/\tau)}{\sum_{b=1}^B \exp(\text{sim}(T_{y_i}, I_b)/\tau)}
\label{eqn:t2isupcon}
\end{equation}

Then, the total loss is given by

\begin{equation}
    \mathcal{L}_{total} = \mathcal{L}_{ID} + \mathcal{L}_{SupCon} 
\label{eqn:totalloss}
\end{equation}

During testing, we use visual feature from the image encoder (after batch normalization applied) $I_i$, and then compare feature of each query image with gallery features using cosine distance.

\noindent

\section{Experimental Results}  \label{sec:experimentalResults}

\subsection{Experimental Setup}

\begin{figure}[!h] 
\begin{minipage}[b]{1.0\linewidth}
  \centering
  \centerline{\includegraphics[width=8.5cm]{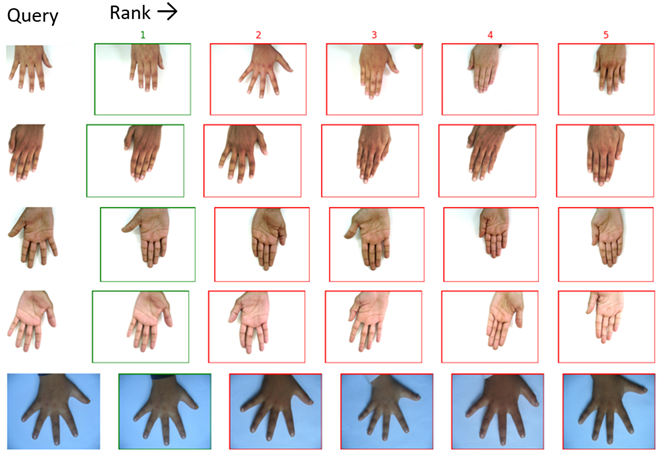}}
\end{minipage}
\caption{Some qualitative results of our method using query vs ranked results retrieved from gallery. From top to bottom row are right dorsal of 11k, left dorsal of 11k, right palmar of 11k, left palmar of 11k and HD datasets. The green and red bounding boxes denote the correct and the wrong matches, respectively.}
\label{fig:result}
\end{figure}

\textbf{Datasets:} We evaluate our approach using two benchmark datasets: the 11k Hands\footnote{\url{https://sites.google.com/view/11khands}} dataset~\cite{Mah19} containing 190 subjects and the Hong Kong Polytechnic University Hand Dorsal (HD)\footnote{\url{http://www4.comp.polyu.edu.hk/~csajaykr/knuckleV2.htm}} dataset~\cite{KumXu16} with 502 identities. Following the partitioning strategy of~\cite{BaiWilRahIPTA22}, we divide the 11k Hands dataset into four distinct subsets: right dorsal (143 identities), left dorsal (146), right palmar (143), and left palmar (151), excluding images with accessories. For both datasets, we split the identities into equal halves for training and testing. The 11k subsets are divided as follows: right dorsal (72 train/71 test), left dorsal (73/73), right palmar (72/71), and left palmar (76/75). Similarly, the HD dataset splits into 251 identities each for training and testing. Our evaluation protocol creates separate gallery and query sets: for each test identity, we randomly select one image for the gallery and use the remaining as queries. The 11k gallery contains 290 images (combined from all subsets), with query sizes of 971 (right dorsal), 988 (left dorsal), 917 (right palmar), and 948 (left palmar). The HD gallery includes 1,593 images (adding 213 low-quality samples without clear knuckle patterns), with 1,992 query images. For validation, we randomly select one image per training identity to monitor training progress. We repeat this entire process 10 times - 10 Monte Carlo query-gallery splits - and report average performance. Fig.~\ref{fig:result} illustrates sample queries from each subset with their corresponding gallery retrieval results.

\noindent \textbf{Implementation details:} Our implementation of CLIP-HandID was developed using the PyTorch framework, with training performed on an NVIDIA GeForce RTX 2080 Ti GPU. As a backbone, we use both ResNet-50 and ViT-B/16 pre-trained from CLIP as our image encoder and a pre-trained text encoder (CLIP text Transformer). In addition, the CLIP-HandID incorporates the inversion network which is random-initialized. During training, input images were first resized to $256 \times 256$ pixels, then randomly cropped to $224 \times 224$. We applied several augmentation techniques including random horizontal flip, normalization, and color jittering. To mitigate over-fitting and excessive confidence in predictions, we employed label smoothing~\cite{SzeVanIof16} with a smoothing factor ($\epsilon$) of 0.1. For testing, images were simply resized to $224 \times 224$ and normalized, without random cropping or other augmentation. In both training and testing phases, we randomized images order through dataset reshuffling. The model was trained for 70 epochs with a batch size of 20 for ViT-B/16 and 4 for ResNet-50 backbones, using cross-entropy and supervised contrastive losses and the Adam optimizer. We applied L2 regularization with a weight decay factor of $5 \times 10^{-4}$. A learning rate warmup strategy~\cite{LuoGuLia19} was implemented for the first 10 epochs, gradually increasing the rate from $5 \times 10^{-8}$ to $5 \times 10^{-6}$. Subsequent learning rate adjustments reduced it to $2.5 \times 10^{-6}$ at epoch 40 and $1.25 \times 10^{-6}$ at epoch 60. Notably, we assigned different learning rates to various network components: the backbone network's existing layers used one-tenth the learning rate of newly added layers (inversion network, FC or linear layers and batch normalizations), with proper initialization of weights and biases.

\subsection{Performance Evaluation}

In this section, we evaluate our model and report the results using mean Average Precision (mAP) and Cumulative Matching Characteristics (CMC)~\cite{ZheSheTia15} (Rank-1 matching accuracy) on 11k hands~\cite{Mah19} and HD~\cite{KumXu16} datasets. Since CLIP supports both CNN-based and ViT-based models for the image encoder, the proposed method is validated on both ResNet-50 and ViT-B/16 backbones of CLIP pre-trained model. We consider three different scenarios for our experiments: zero-shot solution, fine-tuning only the image encoder, and our proposed method, CLIP-HandID, where the inversion network and frozen text encoder are used during training. In all of these scenarios, the text encoder is frozen. In the zero-shot evaluations, both the image encoder and the text encoder are frozen.

As shown in Table~\ref{tbl:sameDomain}, fine-tuning only the image encoder (CLIP-FT-RN50 and CLIP-FT-ViT-B-16) increases the performance over the zero-shot solution (CLIP-ZS-RN50 and CLIP-ZS-ViT-B-16) by a large margin in both rank-1 accuracy and mAP on all datasets. For instance, the fine-tuning of only the image encoder using ResNet-50 backbone increases from 65.80\% to 93.80\% rank-1 accuracy (28.00\% increment) and from 69.11\% to 95.00\% mAP (25.89\% increment) on right dorsal (D-r) of 11k dataset (CLIP-FT-RN50 vs CLIP-ZS-RN50 in Table~\ref{tbl:sameDomain}). Our proposed method which uses the inversion network also increases performance over fine-tuning only the image encoder. For instance, using ViT-B-16 backbone, our proposed method increases the performance from 93.60\% to 95.89\% rank-1 accuracy (2.29\% increment) and from 94.36\% to 96.54\% mAP (2.18\% increment) on left palmar (P-l) of 11k dataset (CLIP-HandID-ViT-B-16 vs CLIP-FT-ViT-B-16 in Table~\ref{tbl:sameDomain}).

The performance of our proposed method is very consistent in outperforming the image encoder only fine-tuning when using ViT-B-16 backbone (CLIP-HandID-ViT-B-16) when compared to using ResNet-50 backbone (CLIP-HandID-RN50). For instance, while CLIP-HandID-ViT-B-16 outperforms CLIP-FT-ViT-B-16 on all datasets, CLIP-HandID-RN50 outperforms CLIP-FT-RN50 only on right and left dorsal of 11k dataset, as shown in Table~\ref{tbl:sameDomain}.

Our proposed method using ViT-B-16 backbone, CLIP-HandID-ViT-B-16, outperforms GPA-Net~\cite{BaiWilRahIPTA22} significantly in both rank-1 accuracy and mAP on all datasets. For instance, our method outperforms the GPA-Net by 1.88\% rank-1 accuracy and by 1.44\% mAP on left dorsal (D-l) of 11k dataset, as shown in Table~\ref{tbl:sameDomain}.

\begin{table*} [htbp]
\caption{\normalfont{Performance comparison of our method with other methods on right dorsal (D-r) of 11k, left dorsal (D-l) of 11k, right palmar (P-r) of 11k, left palmar (P-l) of 11k and HD datasets. Rank-1 accuracy (\%) and mAP (\%) are shown. Best and second best results are shown in $\color{red}{\textbf{red}}$ and $\color{blue}{\textbf{blue}}$, respectively.}}
\label{tbl:sameDomain}
\begin{center}
  \begin{tabular}{|l|ll|ll|ll|ll|ll|}
    \hline
    \multirow{2}{*}{Method} & 
      \multicolumn{2}{c|}{D-r of 11k} & 
      \multicolumn{2}{c|}{D-l of 11k} & 
      \multicolumn{2}{c|}{P-r of 11k} &
      \multicolumn{2}{c|}{P-l of 11k} &
      \multicolumn{2}{c|}{HD} \\
      \cline{2-11}
    & rank-1 & mAP & rank-1 & mAP & rank-1 & mAP & rank-1 & mAP & rank-1 & mAP \\
    \hline
    GPA-Net~\cite{BaiWilRahIPTA22} & 94.80 & 95.72 & 94.87 & 95.93 & \textcolor{blue}{\textbf{95.83}} & \textcolor{blue}{\textbf{96.31}} & \textcolor{blue}{\textbf{95.72}} & \textcolor{blue}{\textbf{96.20}} & 94.64 & 95.08 \\
    \hline
    CLIP-ZS-RN50 & 65.80 & 69.11 & 67.98 & 71.22 & 65.19 & 68.53 & 60.74 & 64.04 & 81.55 & 83.83\\
    CLIP-FT-RN50 & 93.80 & 95.00 & 95.58 & 96.36 & 93.10 & 94.22 & 93.55 & 94.34 & 94.05 & 94.47 \\  
    \textbf{CLIP-HandID}-RN50 (ours) & \textcolor{blue}{\textbf{94.88}} & \textcolor{blue}{\textbf{95.92}} & 95.91 & 96.63 & 91.95 & 93.21 & 92.72 & 93.69 & 93.74 & 94.15 \\
    \hline
    CLIP-ZS-ViT-B-16 & 70.44 & 74.31 & 75.47 & 78.59 & 59.90 & 63.91 & 62.24 & 65.98 & 81.23 & 83.51\\
     CLIP-FT-ViT-B-16 & 94.03 & 95.28 & \textcolor{blue}{\textbf{96.46}} & \textcolor{blue}{\textbf{97.25}} & 95.43 & 96.28 & 93.60 & 94.36 & \textcolor{blue}{\textbf{95.11}} & \textcolor{blue}{\textbf{95.47}} \\ 
    \textbf{CLIP-HandID}-ViT-B-16 (ours) & \textcolor{red}{\textbf{95.68}} & \textcolor{red}{\textbf{96.53}} & \textcolor{red}{\textbf{96.75}} & \textcolor{red}{\textbf{97.37}} & \textcolor{red}{\textbf{97.51}} & \textcolor{red}{\textbf{97.96}} & \textcolor{red}{\textbf{95.89}} & \textcolor{red}{\textbf{96.54}} & \textcolor{red}{\textbf{95.51}} & \textcolor{red}{\textbf{95.87}} \\ 
    \hline
  \end{tabular}
\end{center}
\end{table*}
\noindent

\section{Conclusion} \label{sec:Conclusion}

In this work, we propose a new hand-based person identification framework tailored for forensic applications, where hand imagery often serves as the primary evidence in severe criminal cases such as sexual assault. Our proposed method, CLIP-HandID, innovatively adapts the CLIP vision-language model to this domain by developing a mechanism to bridge the gap between numeric identity labels and semantic visual representations. The core innovation involves learning task-specific pseudo-tokens through textual inversion, which encode distinctive hand attributes (e.g., vein patterns, knuckle geometry, etc.) into adaptable text prompts. These learned pseudo-tokens are integrated into structured semantic prompts (e.g., "A photo of a $S_*$ hand", where
$S_*$ is the learnt pseudo-token) that guide CLIP's cross-modal alignment. The image encoder processes input hand images while the text encoder interprets our enhanced prompts, enabling more discriminative feature learning through the model's pre-trained multi-modal reasoning capabilities. Comprehensive experiments on two large-scale, ethnically diverse hand datasets demonstrate that CLIP-HandID significantly outperforms current state-of-the-art methods in recognition accuracy. This advancement is particularly crucial for forensic investigations where conventional biometric features may be unavailable.


\bibliographystyle{IEEEtran}
\bibliography{refs}

\end{document}